# Boltzmann Classifier: A Thermodynamic-Inspired Approach to Supervised Learning

Muhamed Amin[1,2]*, Bernard R. Brooks[1]

[1]*Laboratory of Computational Biology, National Heart, Lung and Blood, Institute, National Institutes of Health, Bethesda, Maryland 20892, USA.*
[2]Department of Sciences, University College Groningen, University of Groningen, 9718 BG Groningen

Email: muhamed.amin@nih.gov

**Abstract:** We present the Boltzmann classifier, a novel distance-based probabilistic classification algorithm inspired by the Boltzmann distribution. Unlike traditional classifiers that produce hard decisions or uncalibrated probabilities, the Boltzmann classifier assigns class probabilities based on the average distance to the nearest neighbors within each class, providing interpretable, physically meaningful outputs. We evaluate the performance of the method across three application domains: molecular activity prediction, oxidation state classification of transition metal complexes, and breast cancer diagnosis. In the molecular activity task, the classifier achieved the highest accuracy in predicting active compounds against two protein targets, with strong correlations observed between the predicted probabilities and experimental $pIC_{50}$ values. For metal complexes, the classifier accurately distinguished between oxidation states II and III for Fe, Mn, and Co, using only metal–ligand bond lengths extracted from crystallographic data, and demonstrated high consistency with known chemical trends. In the breast cancer dataset, the classifier achieved 97% accuracy, with low-confidence predictions concentrated in inherently ambiguous cases. Across all tasks, the Boltzmann classifier performed competitively or better than standard models such as logistic regression, support vector machines, random forests, and k-nearest neighbors. Its probabilistic outputs were found to correlate with continuous physical or biological properties, highlighting its potential utility in both classification and regression contexts. The results suggest that the Boltzmann classifier is a robust and interpretable alternative to conventional machine learning approaches, particularly in scientific domains where underlying structure–property relationships are important.

## 1. Introduction

Classification remains a fundamental problem in machine learning, with widespread applications across biomedical, financial, and industrial domains. Traditional classifiers such as logistic regression,[1, 2] support vector machines (SVM),[3, 4] and decision trees[5] offer robust solutions but often lack a direct physical interpretability. Inspired by concepts from statistical physics, we introduce the Boltzmann Classifier, a novel method that estimates class probabilities using an energy-based model derived from the Boltzmann distribution.[6] Lower energy configurations correspond to higher probabilities, capturing the intuition that a sample more similar to a class prototype (in terms of feature deviations) is more likely to belong to that class. For example, Boltzmann distribution is used to classify the protonation states of amino acids in proteins based on an electrostatic energy function.[7, 8] This paper presents the theoretical foundation, implementation details, and experimental evaluation of the Boltzmann Classifier using a well-known biomedical dataset.

The Boltzmann distribution, a cornerstone of statistical mechanics, describes the probability of a system occupying a state with energy at temperature as, where is the Boltzmann constant. In machine learning, energy-based models (EBMs) such as Boltzmann machines and Hopfield networks also exploit energy minimization principles to guide learning and inference.[9-12] Our work adapts this concept for supervised classification, using feature deviations to define a simple, interpretable energy function. This energy function is inspired by the bond stretching and the torsion energies in molecular dynamics forcefields, which the energy is proportional for the deviation of the bond length/bond angle from the equilibrium value.[13, 14] Unlike deep EBMs,[15-18] our classifier does not require iterative optimization or backpropagation and is thus computationally efficient and easy to integrate into existing workflows.

One of the primary advantages of the Boltzmann distribution is its ability to generate a meaningful probability distribution across the set of possible states.[19] Moreover, by modulating the temperature parameter, the accessibility of high-energy states can be increased. These properties are particularly advantageous in classification tasks, as the probability outputs provide a more granular level of information compared to a binary or categorical classification. Additionally, the temperature parameter—acting as a global variable—can be leveraged to adjust the likelihood of lower-energy states, which could be beneficial in domains such as security. For instance, the classifier's decision threshold could be tuned to be more stringent, improving its ability to discriminate between legitimate and spam emails.

Here, we introduce the Boltzmann Classifier, unlike traditional classifiers based on softmax or nearest neighbor approaches, the Boltzmann classifier introduces a fundamentally different mechanism for estimating class probabilities. In softmax-based models, class scores (logits) are computed via linear or nonlinear transformations of the input, and probabilities are derived through normalization over these scores. These models typically require gradient-based training and rely on parameterized representations. In contrast, the Boltzmann classifier is a non-parametric, distance-based model that estimates the probability of class membership using an energy formulation inspired by statistical mechanics. Specifically, it computes an energy score for each class based on the average distance between a test point and its n nearest neighbors in that class and then converts these energy scores into probabilities using a Boltzmann distribution. While conceptually related to k-nearest neighbor (k-NN) classifiers, which assign labels based on majority voting among local neighbors, the Boltzmann classifier goes further by producing smooth, probabilistic outputs that reflect both local structure and class separability. This combination of interpretability, uncertainty quantification, and model-free design makes the Boltzmann classifier an appealing alternative for applications where geometric relationships in feature space carry physical or biological meaning.

**Computational Methods**

The Boltzmann Classifier estimates the probability of each class given an input vector x by comparing it to the mean feature vector $\mu_c$ of each class, computed from the training data. The energy of assigning to class is defined as the L1-norm deviation:

$$E_c(\mathbf{x}) = \frac{1}{n} \sum_{\mathbf{x}_i \in N_n^{(c)}(\mathbf{x})} d(\mathbf{x}, \mathbf{x}_i)$$

Where $d(x, x_i)$ is the Euclidean distance, $N_n^{(c)}(x)$ is the set of the n nearest neighbors of x in class c. The class probabilities are then calculated as:

$$P(c|x) = \frac{e^{-E_c(x)/kT}}{\sum_j e^{-E_j(x)/kT}}$$

Where is a scaling factor analogous to the Boltzmann constant and controls the softness of the distribution. The class with the highest probability is selected as the predicted label. All features are preprocessed using MinMax scaling to ensure consistent contributions to the energy function.

*Implementation.* We implemented the Boltzmann Classifier in Python using a class that inherits from `BaseEstimator and ClassifierMixin` from the scikit-learn library.[20] This design ensures compatibility with pipelines, cross-validation tools, and hyperparameter tuning utilities provided by scikit-learn. The `fit` method computes the mean feature vector for each class in the training set, while the `predict` method calculates the energy for each class and applies the Boltzmann probability formula to determine the predicted label. Feature scaling is performed using `MinMaxScaler` to normalize all features to the range [0, 1].

The energy function currently uses the L1 norm (absolute deviation), but this can be generalized to other distance measures. The implementation is compact, computationally efficient, and interpretable, making it an attractive alternative to black-box models.

## Results and discussion

To evaluate the performance and interpretability of the proposed Boltzmann classifier, we conducted a series of experiments across three distinct application domains. These case studies were designed to test the classifier's ability to handle real-world data from drug discovery, chemical informatics, and biomedical domains, each with unique structural and statistical characteristics. Specifically, we assessed: (1) molecular activity prediction for drug-like compounds against two protein targets; (2) oxidation state classification of transition metals based on crystallographic bond lengths; and (3) diagnosis of breast cancer using clinical features. In each case, the Boltzmann classifier was benchmarked against several established machine learning algorithms, including Logistic Regression, Support Vector Machines, Random Forest, and k-Nearest Neighbors, using consistent evaluation protocols. In addition to accuracy, we analyzed the correlation between the classifier's probabilistic outputs and underlying physical or biological variables to assess interpretability and potential regression capability. The following sections present the results and insights from each application domain.

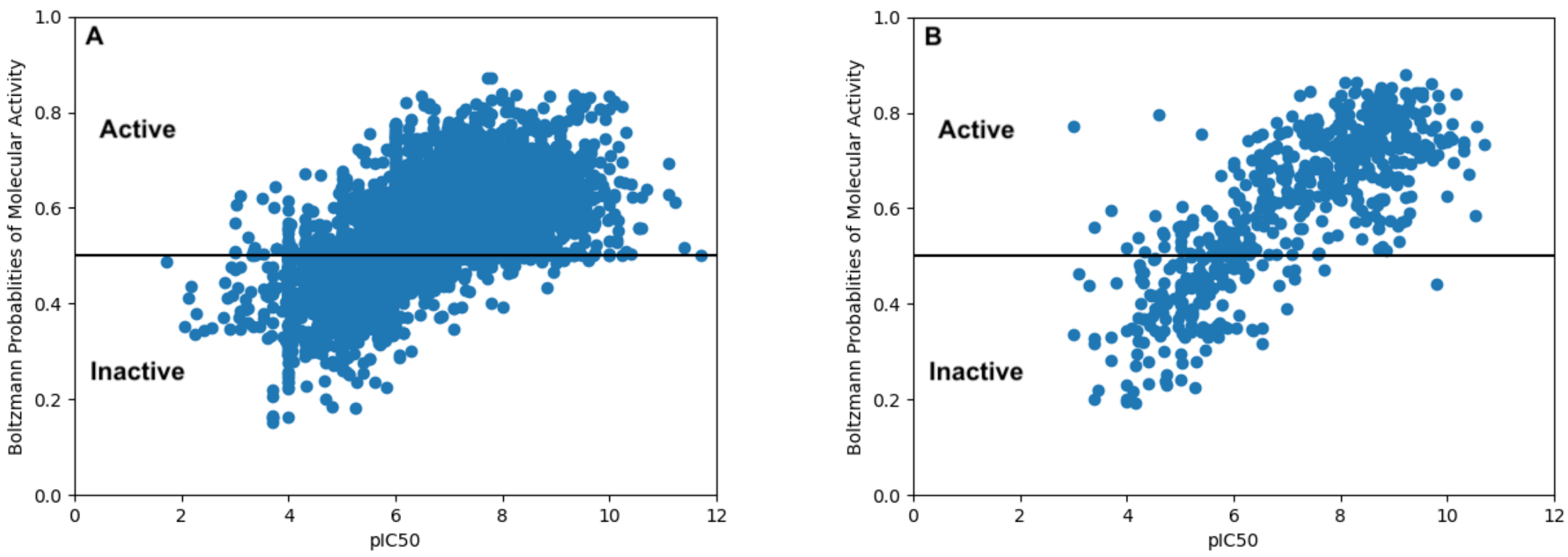

**Figure 1**. The correlation between the Boltzmann probabilities of the molecular activities and the pIC50 of A) CHEMBL203 with $R^2$ is 0.61. B) CHEMBL1827 with $R^2$ is 0.78.

**Table 1.** The accuracy score of the different ML classifiers for molecular activity classification.

| | CHEMBL203 | CHEMBL1827 |
|---|---|---|
| Boltzmann Classifier | 0.85 | 0.89 |
| Logistic Regression | 0.80 | 0.86 |
| Support Vector Machines | 0.81 | 0.85 |
| Random Forest | 0.85 | 0.89 |
| Nearest Neighbor | 0.83 | 0.87 |

The BC and the RF achieved the highest accuracy scores for the two target proteins.

*Molecular activity prediction*. To assess the effectiveness of the Boltzmann classifier in molecular activity prediction, we applied it to two well-characterized protein targets: CHEMBL203 (Epidermal Growth Factor Receptor, EGFR) and CHEMBL1827 (Tyrosine-protein kinase ABL1).[21] Compounds were labeled as active if their experimentally determined $pIC_{50}$ values exceeded 6.0 (corresponding to an $IC_{50} < 1$ μM), and as inactive otherwise. The datasets were obtained from the ChEMBL database and included 16,693 compounds for CHEMBL203 and 2,414 compounds for CHEMBL1827.

Molecular representations were computed using RDKit,[22] from which all 237 physicochemical and topological descriptors were extracted for each compound. The datasets were randomly split into training and test sets, with 25% of the data reserved for testing in each case.

We compared the classification performance of the Boltzmann classifier against four widely used machine learning models: Logistic Regression, Support Vector Machines (SVM), Random Forest, and k-Nearest Neighbors (k-NN). Across both protein targets, the Boltzmann classifier achieved the highest accuracy, matching the performance of the Random Forest classifier, and outperforming the other methods (Table 1).

In addition to classification performance, we investigated the relationship between the predicted Boltzmann probabilities for the active class and the corresponding experimental $pIC_{50}$ values. A significant positive correlation was observed for both protein targets: Pearson correlation coefficients of 0.61 for CHEMBL203 and 0.78 for CHEMBL1827, indicating that higher predicted probabilities are associated with higher biological activity. These findings suggest that the Boltzmann classifier, despite being trained in a purely categorical setting, inherently captures quantitative trends in molecular potency. This opens the possibility of using the Boltzmann framework for regression tasks or for prioritizing compounds based on predicted activity scores. Corresponding scatter plots illustrating the relationship between predicted probabilities and $pIC_{50}$ values are included for both targets.

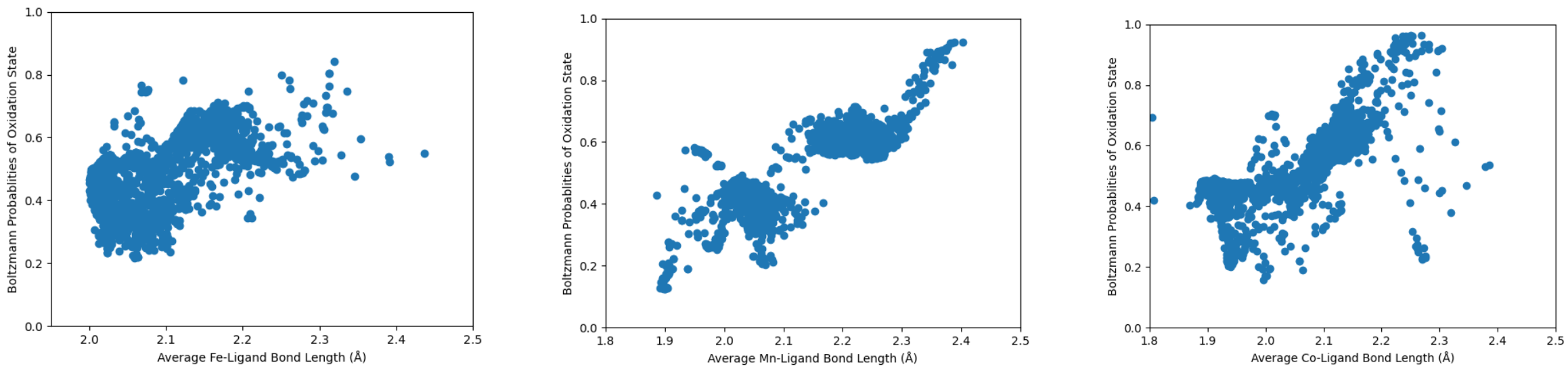

**Figure 2**. A) The correlation between the Boltzmann probabilities of the Fe oxidation state and the average bond length. The $R^2$ is 0.50. B) The correlation between the Boltzmann probabilities of the Mn oxidation and the average bond length. The $R^2$ is 0.85. C) The correlation between the Boltzmann probabilities of the Co oxidation and the average bond length. The $R^2$ is 0.68.

**Table 2.** The accuracy score of the different ML classifiers of metal oxidation states classification.

| | Fe | Mn | Co |
|---|---|---|---|
| Boltzmann Classifier | 0.82 | 0.81 | 0.89 |
| Logistic Regression | 0.77 | 0.81 | 0.85 |
| Support Vector Machines | 0.82 | 0.81 | 0.87 |
| Random Forest | 0.81 | 0.70 | 0.85 |
| Nearest Neighbor | 0.79 | 0.81 | 0.86 |

The BC achieved the highest accuracy scores for the three metals.

*Oxidation State Classification in Transition Metal Complexes*. As a second evaluation, the Boltzmann classifier was applied to the prediction of oxidation states (II and III) for transition metal centers in octahedral complexes, using structural data extracted from the Cambridge Crystallographic Data Centre (CCDC).[23] The study focused on three transition metals: iron (Fe), manganese (Mn), and cobalt (Co), comprising 9,360, 8,155, and 9,375 structures respectively. A custom Python script was developed to extract the six bond lengths corresponding to the metal–ligand distances in each octahedral environment, which were used as input features for classification.

The Boltzmann classifier was compared against standard machine learning models including Logistic Regression, Support Vector Machines (SVM), Random Forest, and k-Nearest Neighbors (k-NN). As shown in Table 2, the Boltzmann classifier achieved the highest accuracy for Mn and Co and matched the top performance of SVM for Fe. Specifically, the Boltzmann classifier reached accuracies of 0.82 (Fe), 0.81 (Mn), and 0.89 (Co). These results demonstrate the classifier's ability to capture subtle structural features indicative of oxidation state and confirm its competitive performance relative to widely used classification algorithms.

To further explore the interpretability of the Boltzmann classifier outputs, we examined the relationship between the predicted Boltzmann probabilities for the higher oxidation state (III) and the average metal–ligand bond length in each complex. As shown in Figure 2, scatter plots of the Boltzmann probabilities versus average bond lengths reveal meaningful trends across all three metal types. We observed Pearson correlation coefficients of 0.50 for Fe, 0.85 for Mn, and 0.68 for Co, indicating that the predicted probabilities are positively correlated with bond length contraction, a well-known structural consequence of higher oxidation states. These results suggest that the Boltzmann probabilities do not merely serve as classification outputs, but also encode chemically relevant structural information, reinforcing the physical interpretability of the model.

*The Breast Cancer Wisconsin dataset.*[24] The dataset consists of 569 samples and 30 numeric features extracted from digitized images of breast masses. Each sample is labeled as either malignant or benign. The Boltzmann classifier was also evaluated on the Wisconsin Breast Cancer dataset, a well-established benchmark for binary classification of malignant and benign tumors. Using the standard train-test split, the classifier achieved an accuracy of 97%, correctly identifying most cases. The confusion matrix revealed only four misclassified samples, indicating a high degree of reliability. Notably, as shown in Table X, the Boltzmann probabilities for these misclassified cases were relatively close to the decision boundary (i.e., near 0.5), suggesting that these instances are inherently ambiguous and difficult to classify, even for other models. This highlights the classifier's ability to express uncertainty in its predictions, providing probabilistic outputs that are not only accurate but also informative for decision support in clinical applications.

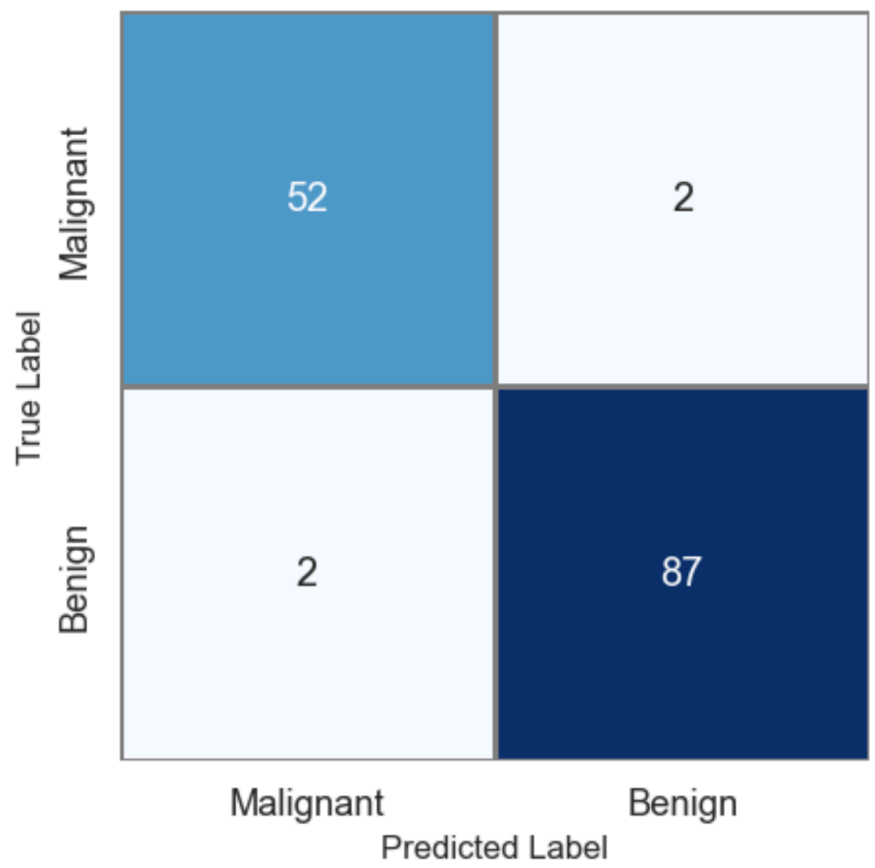

**Table 1**. The probabilities of misclassified elements in the breast cancer dataset

| Malignant | Benign |
|---|---|
| 0.525 | 0.475 |
| 0.477 | 0.523 |
| 0.505 | 0.495 |
| 0.482 | 0.518 |

The probabilities are almost equal for all misclassified elements.

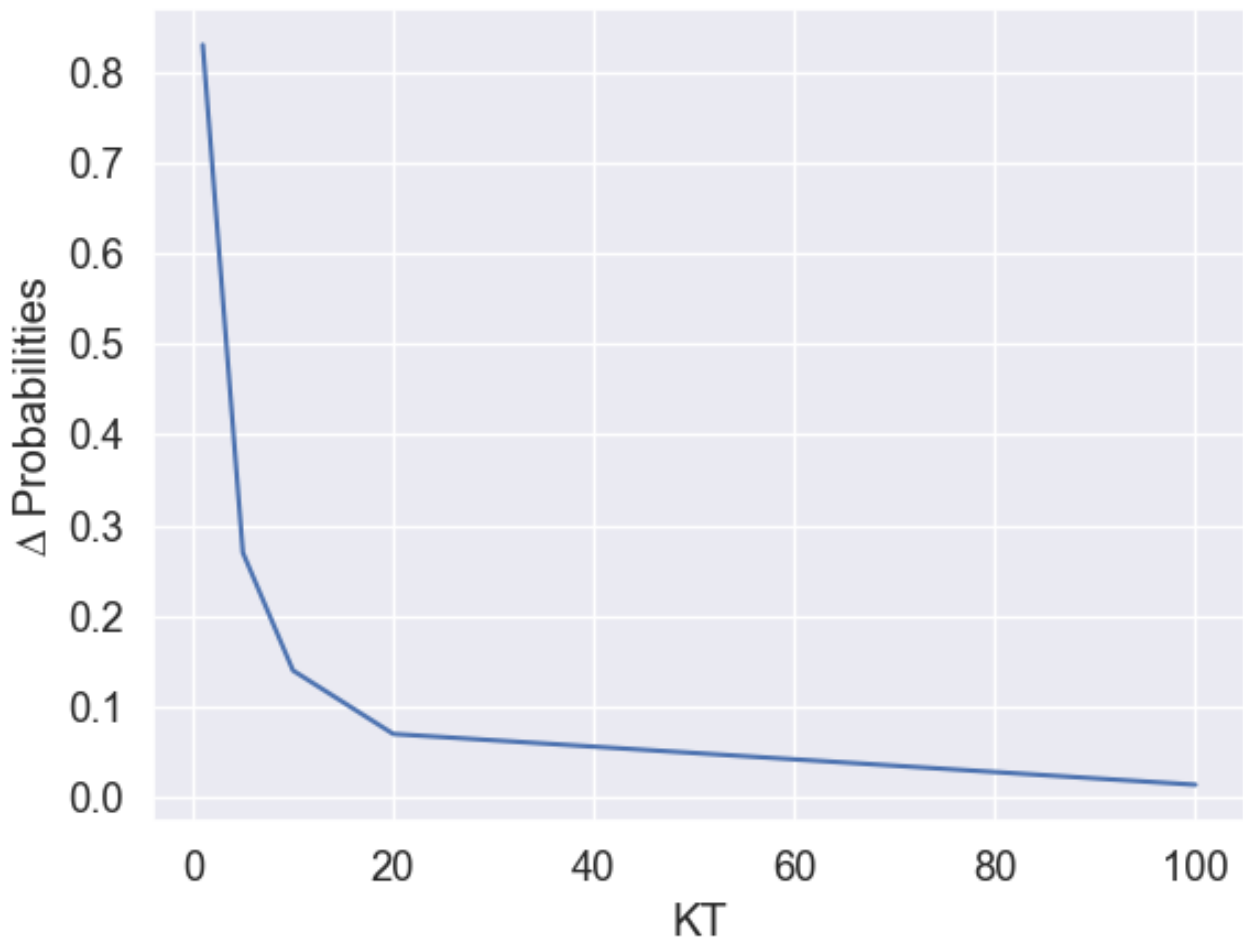

**Figure 3**. The average Δ probabilities (p(malignant)-p(benign)) for the correctly predicted classes vs. the KT variable.

As an analogue to microstate probabilities in Boltzmann distribution, raising the temperature allows the high energy states to be more occupied. Thus, we studied the effect of increasing KT on the average Δ probabilities between the two classes (malignant and benign) for the correctly assigned instances. As shown in Figure 3, increasing the KT reduces the Δ probabilities as expected, which means that the high energy states are more occupied at higher KT values.

**Conclusion**

In this study, we introduced and evaluated the Boltzmann classifier, a distance-based probabilistic classification algorithm inspired by the Boltzmann distribution. Across three diverse and domain-relevant applications—molecular activity prediction, oxidation state classification of transition metal complexes, and breast cancer diagnosis—the Boltzmann classifier consistently demonstrated strong predictive performance. It achieved classification accuracies comparable to or exceeding those of standard machine learning models such as logistic regression, support vector machines, random forest, and k-nearest neighbors. Beyond accuracy, the classifier provided interpretable probabilistic outputs that were found to correlate meaningfully with underlying physical or biological variables, including $pIC_{50}$ values and average bond lengths. These correlations suggest that the Boltzmann classifier not only excels in classification tasks but also captures underlying structure–activity and structure–property relationships, making it a promising tool for applications requiring both predictive power and interpretability.

A potential limitation of the Boltzmann classifier is its reliance on distance metrics and nearest-neighbor computations, which can become computationally expensive for large, high-dimensional datasets. Moreover, its performance may be sensitive to feature scaling and the choice of distance function. These aspects warrant further optimization and investigation, particularly in complex, noisy, or unbalanced datasets. Future work may explore extensions to regression settings, integration with feature learning techniques (e.g., autoencoders or graph neural networks), and its potential synergy with energy-based models for improved scalability and representation learning.

**Code availability**

The implementation of Boltzmann Classifier within the scikit-learn framework is available in the following GitHub repository: https://github.com/mamin03/BoltzmannClassifier.git

**Author Contribution**

Muhamed Amin: Conceptualization, code implementation, writing, analysis.
Bernard R. Brooks: Writing, analysis, funding